\documentclass[10pt,twocolumn,letterpaper]{article}

\usepackage{iccv}              

\definecolor{iccvblue}{rgb}{0.21,0.49,0.74}
\usepackage[pagebackref,breaklinks,colorlinks,allcolors=iccvblue]{hyperref}

\usepackage{hyperref}
\hypersetup{
colorlinks = true, 
urlcolor = magenta, 
linkcolor = red, 
citecolor = blue}
\usepackage{multirow}
\usepackage{enumitem}
\usepackage{romannum}
\usepackage{booktabs}
\usepackage{amsmath}
\usepackage{cuted}
\usepackage{capt-of}

\usepackage{array}
\newcolumntype{P}[1]{>{\centering\arraybackslash}p{#1}}

\usepackage{xcolor,colortbl}
\definecolor{Gray}{gray}{0.9}

\def\paperID{5} 
\def\confName{ICCV}
\def\confYear{2025}

\title{DepthFlow: Exploiting Depth-Flow Structural Correlations\\for Unsupervised Video Object Segmentation}

\author{Suhwan Cho$^1$\quad Minhyeok Lee$^2$\quad Jungho Lee$^2$\quad Donghyeong Kim$^2$\quad Sangyoun Lee$^2$\vspace{0.2cm}\\
$^1$~GenGenAI\quad $^2$~Yonsei University\vspace{0.1cm}\\
\fontsize{10.0}{10.0}\url{https://github.com/suhwan-cho/DepthFlow}\\
}

\begin{document}
\maketitle

\begin{abstract}
Unsupervised video object segmentation (VOS) aims to detect the most prominent object in a video. Recently, two-stream approaches that leverage both RGB images and optical flow have gained significant attention, but their performance is fundamentally constrained by the scarcity of training data. To address this, we propose DepthFlow, a novel data generation method that synthesizes optical flow from single images. Our approach is driven by the key insight that VOS models depend more on structural information embedded in flow maps than on their geometric accuracy, and that this structure is highly correlated with depth. We first estimate a depth map from a source image and then convert it into a synthetic flow field that preserves essential structural cues. This process enables the transformation of large-scale image-mask pairs into image-flow-mask training pairs, dramatically expanding the data available for network training. By training a simple encoder-decoder architecture with our synthesized data, we achieve new state-of-the-art performance on all public VOS benchmarks, demonstrating a scalable and effective solution to the data scarcity problem.
\end{abstract}

\begin{figure}[t]
\centering
\includegraphics[width=1\linewidth]{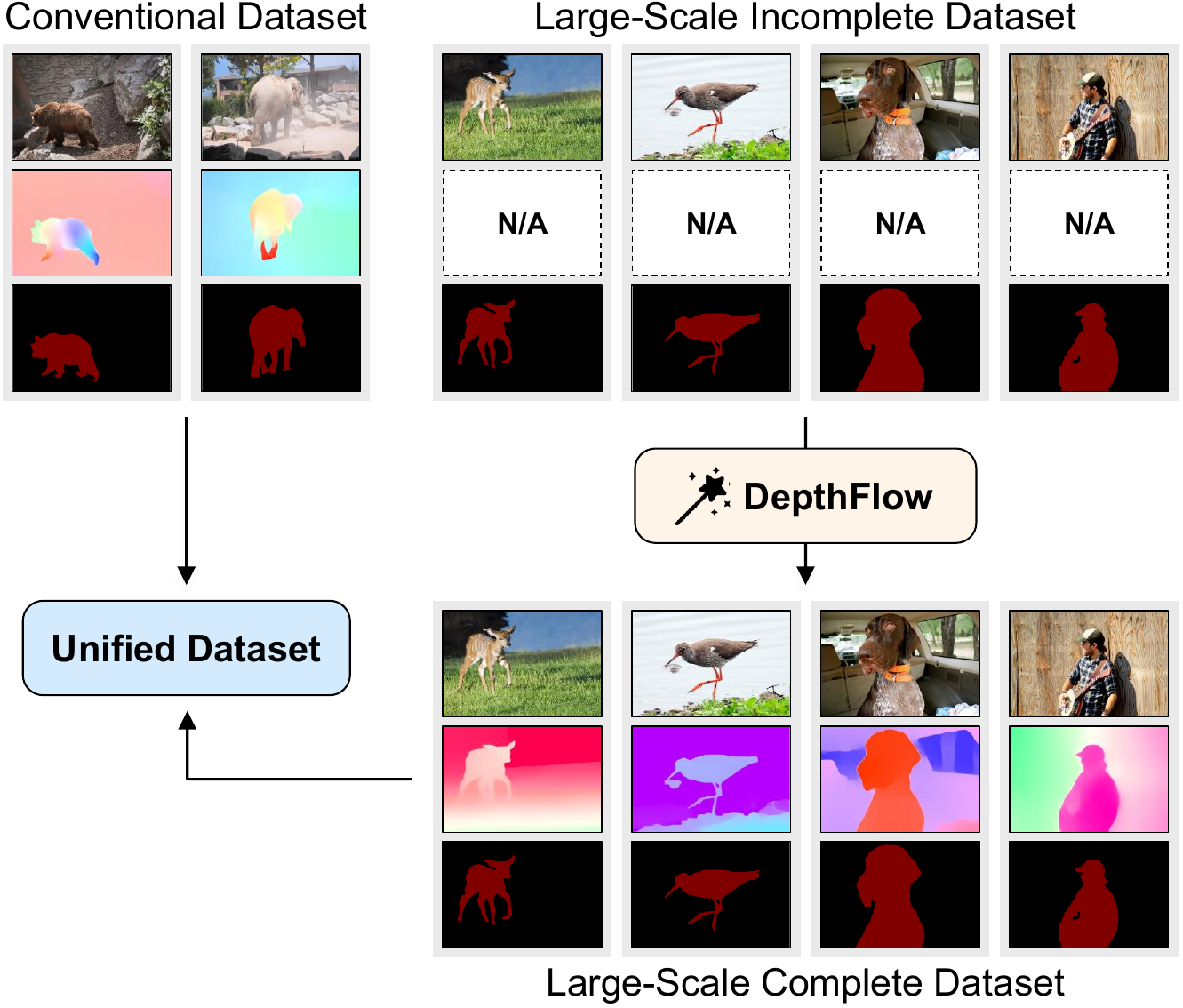}
\vspace{-6mm}
\caption{Overview of our synthetic data generation approach. We convert estimated depth maps from single images into plausible optical flows, creating large-scale image-flow-mask training pairs.}
\label{figure1}
\end{figure}

\section{Introduction}
Video object segmentation (VOS) is a fundamental task in computer vision, playing a crucial role in various applications. Depending on the use case, VOS may incorporate external guidance such as initial frame masks, points, human interaction, or language expressions. This study focuses on the unsupervised setting, where the goal is to segment the most salient object without any external guidance.

In unsupervised VOS, also referred to as video salient object detection (SOD), two-stream approaches have recently gained considerable attention. These methods typically extract optical flow maps between adjacent frames using optical flow estimation models, then leverage image-flow pairs as inputs to generate segmentation masks. Consequently, network training requires complete datasets comprising RGB images, optical flow maps, and segmentation masks. However, their performance is fundamentally constrained by the scarcity of such complete datasets, which remains a significant bottleneck.

Existing methods attempt to address this challenge by leveraging large-scale incomplete datasets, such as general video segmentation datasets~\cite{YTVOS, mose, MeViS} or image-level SOD datasets~\cite{DUTS, MSRA10K, ECSSD}. However, video segmentation datasets often include masks unsuitable for unsupervised VOS (e.g., non-salient objects), while image-level SOD datasets lack optical flow maps, limiting their effectiveness as primary training sources. Consequently, fine-tuning on small-scale complete datasets becomes necessary but remains insufficient for robust model training.

To address this, we propose DepthFlow, a novel data generation method that synthesizes optical flow from single images. Our approach is based on the key insight that VOS models rely on the structural information in flow maps, not their geometric precision, and that this structure is highly correlated with depth. We first estimate a depth map from a source image and then convert it into a synthetic flow field that preserves essential structural motion cues. This process enables the transformation of large-scale image-mask pairs into image-flow-mask training pairs (as shown in Figure~\ref{figure1}), dramatically addressing data scarcity while maintaining the structural information necessary for effective segmentation.

By training a simple encoder-decoder architecture with our synthesized data, we achieve new state-of-the-art performance on all public benchmark datasets, demonstrating a scalable solution to the data scarcity problem.

Our main contributions are summarized as follows:
\begin{itemize}[leftmargin=0.2in]
\item We propose DepthFlow, a novel data generation method that synthesizes optical flow from single images by converting estimated depth maps while preserving structural cues to transform image-mask pairs into large-scale image-flow-mask triplets.

\item We demonstrate that VOS models can be trained effectively with synthetic flows that preserve structural information, achieving state-of-the-art performance using a simple encoder-decoder architecture.

\item Our approach provides a scalable solution to the data scarcity problem by dramatically expanding available training data through image-to-flow conversion.
\end{itemize}

\begin{figure*}[t]
\centering
\includegraphics[width=1\linewidth]{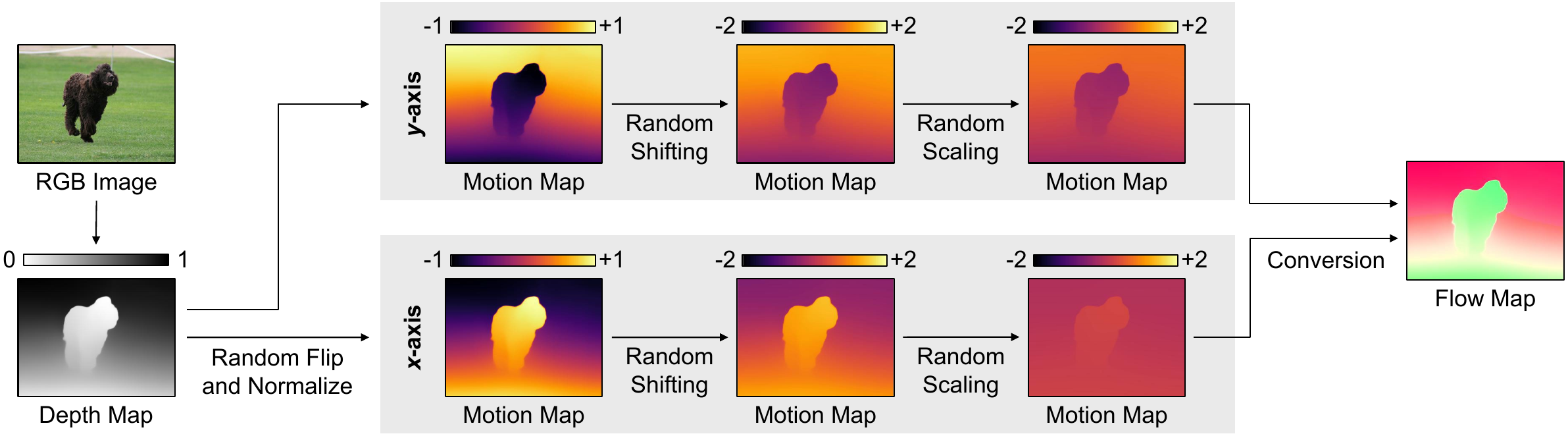}
\vspace{-6mm}
\caption{Illustration of our synthetic flow generation process from estimated depth maps.}
\label{figure2}
\end{figure*}

\section{Related Work}
\noindent\textbf{Temporal reasoning.}
Many unsupervised VOS methods leverage temporal coherence across video frames to improve segmentation consistency. Early approaches like AGS~\cite{AGS} employ recurrent neural networks to model temporal dynamics while preserving spatial details. More recent methods explore various temporal modeling strategies: COSNet~\cite{COSNet} uses co-attention modules for dense frame-pair comparisons, AGNN~\cite{AGNN} treats frames as graph nodes with attentive propagation, and DFNet~\cite{DFNet} models long-term correlations through discriminative feature learning. Reference-based methods include AD-Net~\cite{AD-Net}, which uses initial frames as anchors for pairwise dependencies, and F2Net~\cite{F2Net}, which extends this with point heatmap estimation. Additionally, 3DC-Seg~\cite{3DC-Seg} and D$^2$Conv3D~\cite{D2Conv3D} adopt 3D convolutions for explicit spatio-temporal modeling, while IMP~\cite{IMP} sequentially connects and propagates image-level predictions.

\vspace{1mm}
\noindent\textbf{Multi-modality fusion.}
To address RGB ambiguities, two-stream architectures integrate optical flow with appearance information. Various fusion strategies have been proposed: MATNet~\cite{MATNet} uses deeply interleaved encoders for hierarchical interactions, RTNet~\cite{RTNet} employs reciprocal transformation networks, and FSNet~\cite{FSNet} utilizes full-duplex mutual propagation. Attention-based methods include AMC-Net~\cite{AMC-Net} with co-attention gating and TransportNet~\cite{TransportNet} with optimal transport matching. HFAN~\cite{HFAN} aligns features across multiple embedding layers, while SimulFlow~\cite{SimulFlow} performs joint extraction at the embedding stage.

To improve robustness, some methods incorporate additional techniques. TMO~\cite{TMO} trains motion encoders that can leverage appearance features when flow quality is poor. OAST~\cite{OAST} and DATTT~\cite{DATTT} employ test-time learning for inference refinement, while GFA~\cite{GFA} uses training augmentation to align train-test distributions. However, these methods are fundamentally limited by the scarcity of complete image-flow-mask training datasets.

\vspace{1mm}
\noindent\textbf{Hybrid approaches.}
Recent methods combine temporal reasoning with multi-modal fusion to leverage both long-term consistency and short-term motion cues. PMN~\cite{PMN} maintains an external memory bank for continuously updating appearance and motion features, while GSA-Net~\cite{GSA-Net} and DPA~\cite{DPA} inject global video properties using reference frames, providing the network
with comprehensive knowledge of each video sequence.

\section{Approach}

\subsection{Preliminaries}

\noindent\textbf{Conventional pipeline.}
Recent unsupervised VOS methods predominantly adopt two-stream architectures that leverage both RGB images $I$ and optical flow maps $F$ for feature extraction. Optical flows are typically calculated as:
\begin{align}
&F^i = \Phi(I^i, I^{i+1})~,
\end{align}
where $\Phi(I_1,I_2)$ computes optical flow from source frame $I_1$ to target frame $I_2$, and $i$ denotes the frame index. For the final frame, the target frame is set to the preceding frame. Given $I$ and $F$ for each frame, the segmentation mask $S$ is obtained as:
\begin{align}
&S = \Psi(I, F; \mathrm{w})~,
\end{align}
where $\Psi$ represents the segmentation network and $\mathrm{w}$ its learnable parameters. This process operates per-frame, assuming short-term motion cues provide sufficient information for primary object detection.

\vspace{1mm}
\noindent\textbf{Training data scarcity.}
While two-stream VOS architectures are effective, their performance is fundamentally constrained by the scarcity of complete training datasets containing image-flow-mask triplets. The most commonly used dataset, DAVIS 2016~\cite{DAVIS} training set, contains only 30 training video sequences, which lacks the diversity required for robust network learning.

To mitigate this limitation, existing methods employ two primary strategies. The first involves using large-scale video segmentation datasets such as YouTube-VOS~\cite{YTVOS} and MOSE~\cite{mose}. However, these datasets contain multiple annotated objects per frame, many of which are non-salient. Converting multi-label annotations to binary masks introduces training noise, making these datasets more suitable for pre-training than primary training. The second strategy utilizes image-level saliency datasets such as DUTS~\cite{DUTS} or MSRA10K~\cite{MSRA10K}. Since these datasets lack optical flow information, models cannot be trained in a complete setup, requiring subsequent fine-tuning on small complete datasets.

Both approaches are imperfect solutions that often lead to overfitting on limited complete datasets, ultimately hindering model generalization and performance.

\vspace{1mm}
\noindent\textbf{Key observations.}
Our study is driven by two key observations about existing two-stream VOS approaches. First, most methods operate per-frame without needing to model long-term temporal relationships. This suggests that image-flow-mask pairs may suffice, eliminating the need for full video sequences. Second, VOS models rely on the structural information in flow maps, not their geometric precision, and this structure is highly correlated with depth. The primary objective is to capture structural motion cues that distinguish foreground from background. Therefore, synthetic flows that preserve essential structural information are sufficient for effective unsupervised VOS.

\subsection{Depth-to-Flow Synthetic Data Generation}
Building on our key observations that unsupervised VOS methods operate per-frame and rely on structural information in flow maps rather than geometric precision, we propose DepthFlow, a novel data generation method that synthesizes optical flow maps from single images. Our approach exploits the strong correlation between depth and flow structure: we first estimate a depth map from a source image and then convert it into a synthetic flow field that preserves essential structural cues. Figure~\ref{figure2} illustrates the complete pipeline.

\begin{figure}[t]
\centering
\includegraphics[width=1\linewidth]{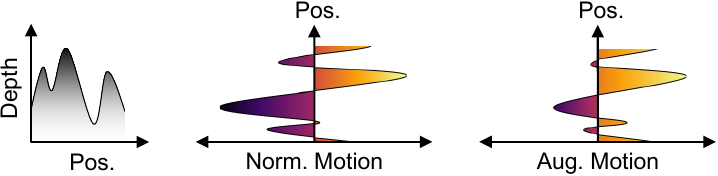}
\vspace{-6mm}
\caption{Visualized depth-to-flow conversion process. Augmented motion indicates the motion map after applying random value shifting and scaling.}
\label{figure3}
\end{figure}

\vspace{1mm}
\noindent\textbf{Depth map estimation.}
We first estimate a depth map from each input image using a pre-trained monocular depth estimation model, DPT-Hybrid~\cite{DPT}. Let $f(\cdot)$ denote the depth estimation function. To ensure consistent value ranges across different images, we apply min-max normalization to obtain $D \in [0,1]$:
\begin{align}
&D = \frac{f(I) - \min(f(I))}{\max(f(I)) - \min(f(I))}~,
\end{align}
where the min and max operations are computed across all pixels in the depth map. This normalization provides a stable foundation for the subsequent flow synthesis process.

\begin{figure*}[t!]
\centering
\includegraphics[width=1\linewidth]{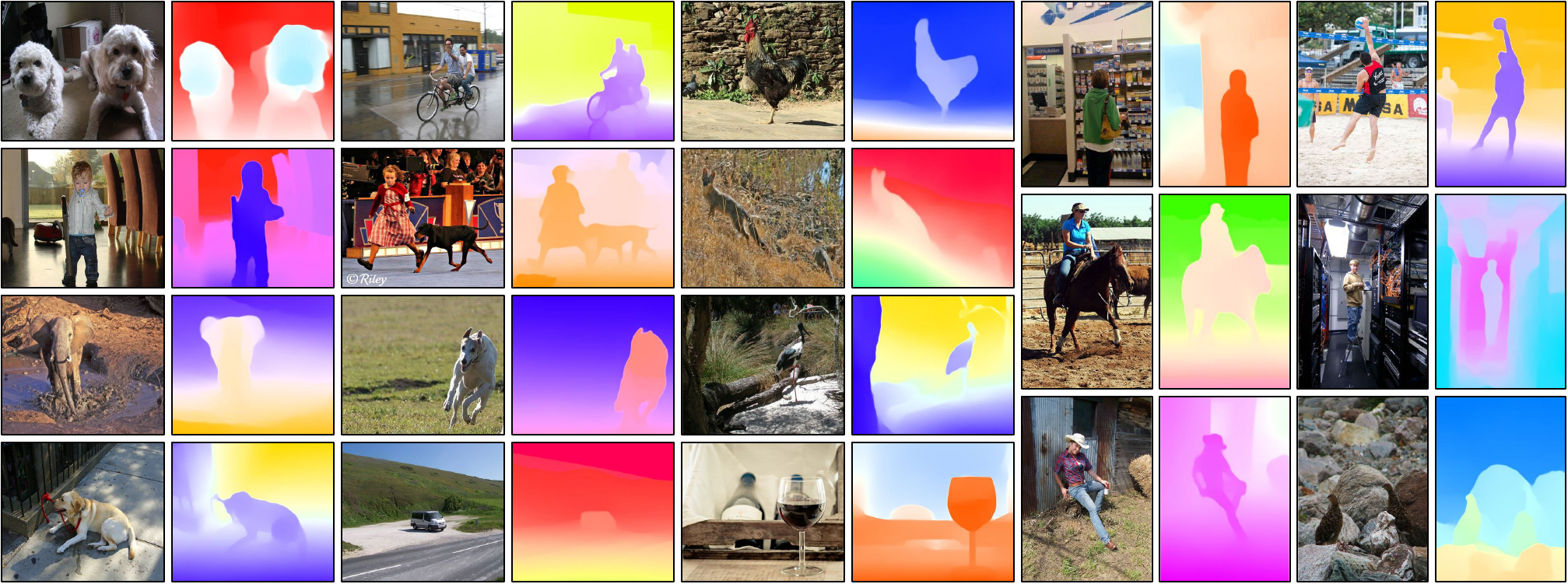}
\vspace{-6mm}
\caption{Representative image-flow pairs from our DUTSv2 dataset.}
\label{figure4}
\end{figure*}

\vspace{1mm}
\noindent\textbf{Motion map generation.}
Given the normalized depth map $D$, we convert it into synthetic optical flow patterns through a series of carefully designed transformations that simulate realistic motion characteristics. The generation process operates independently along the $x$ and $y$ axes, creating diverse two-channel optical flow maps while preserving the spatial relationships inherent in depth structures.

We first apply random depth reversal to create motion diversity with both positive and negative directions:
\begin{align}
&M_1 = 2r(1-D) + 2(1-r)D - 1, \quad r \overset{\mathrm{R}}{\leftarrow} \{0, 1\}~,
\end{align}
where $r$ is a randomly selected binary value. This transformation generates diverse motion patterns by randomly determining the polarity of the depth-to-motion mapping, while normalizing values to the range $[-1, 1]$ to enable both positive and negative motion directions. Next, we introduce random value shifting to further enhance motion diversity:
\begin{align}
&M_2 = M_1 + s, \quad s \overset{\mathrm{R}}{\leftarrow} [-1, 1]~.
\end{align}
By adding random offsets $s$ uniformly across all pixels, we shift the motion baseline to create additional directional variations, enabling comprehensive coverage of both positive and negative pixel displacements that characterize real optical flow patterns. Finally, we apply random value scaling to control motion magnitude and simulate varying flow intensities:
\begin{align}
&M_3 = \alpha \cdot M_2, \quad \alpha \overset{\mathrm{R}}{\leftarrow} [0, 1]~.
\end{align}
This scaling operation modulates the optical flow magnitude from subtle to pronounced movements, reflecting the diverse range of motion intensities encountered in real video sequences. Figure~\ref{figure3} visualizes this motion map generation process, demonstrating how diverse synthetic optical flow patterns are systematically generated from a single depth map through these sequential transformations.

\vspace{1mm}
\noindent\textbf{Flow map visualization.}
The resulting two-channel motion map $M_3 \in [-2, 2]$ requires normalization and conversion to RGB format for compatibility with standard optical flow processing pipelines. We first normalize the motion values by scaling the maximum absolute value in both horizontal and vertical components to 1:
\begin{align}
&M = \frac{M_3}{\max(\max(|M_3^x|), \max(|M_3^y|))}~,
\end{align}
where $M_3^x$ and $M_3^y$ represent the horizontal and vertical components of the motion map, respectively. This normalization preserves relative motion patterns while standardizing the dynamic range for consistent processing.
The normalized motion map is then converted to RGB format using the standard flow visualization protocol:
\begin{align}
&F = \text{UV2RGB}(M)~,
\end{align}
where UV2RGB represents the conventional mapping function that transforms two-channel motion vectors into three-channel RGB flow visualizations suitable for feature embedding in optical flow processing networks.

\begin{table}[t!]
\centering 
\small
\caption{Training data comparison showing the substantial increase in scale and diversity achieved by our synthetic data.}
\vspace{-2mm}
\begin{tabular}{c|c|c}
\toprule
Dataset &\#Triplets &\#Visual Contexts\\
\midrule
Real &2,079 &30\\
Synthetic &15,572 &15,572\\
Mixed &17,651 &15,602\\
\bottomrule
\end{tabular}
\label{table1}
\end{table}

\begin{figure*}[t]
\centering
\includegraphics[width=1\linewidth]{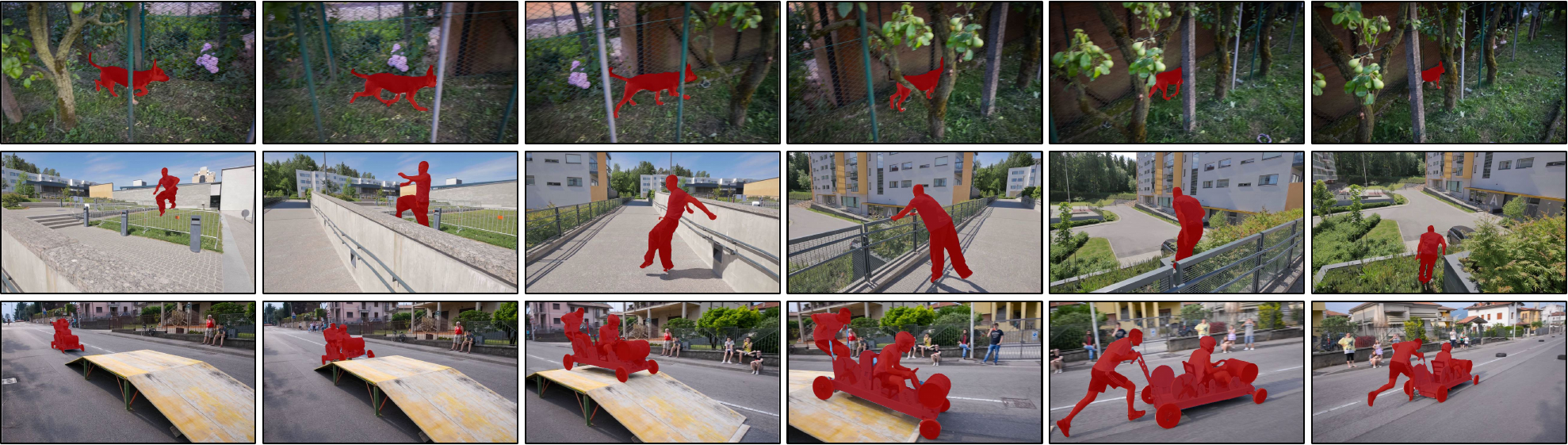}
\vspace{-6mm}
\caption{Qualitative results on the DAVIS 2016 validation set.}
\label{figure5}
\end{figure*}

\subsection{Large-Scale Dataset Construction}
We apply our depth-to-flow synthetic data generation method to construct a comprehensive training dataset for two-stream VOS architectures. Using DUTS~\cite{DUTS}, a large-scale image-level SOD dataset containing 10,553 training and 5,019 testing image-mask pairs, as our foundation, we generate synthetic flow maps for all images in both splits. The resulting image-flow-mask triplets form our augmented dataset, referred to as DUTSv2, which provides over 15,000 complete training samples compared to the 30 video sequences available in DAVIS 2016~\cite{DAVIS}. Figure~\ref{figure4} showcases representative image-flow pairs from DUTSv2, demonstrating the diversity and plausibility of our synthetic flows across various scene types and object categories.

Table~\ref{table1} provides a quantitative comparison of training data scale and diversity between real and synthetic datasets. Here, visual diversity is defined as the number of distinct scenes or visual environments used during training. While real video dataset, DAVIS 2016 training set, contains multiple frames from the same 30 video sequences (resulting in 2,079 training pairs but only 30 distinct visual environments), our synthetic approach generates flows from individual images, providing 15,572 unique visual environments. This dramatic increase in visual diversity is crucial for network generalization, as it exposes the model to a much broader range of scenes, objects, and motion patterns than real video-based training. The mixed dataset combines both real and synthetic data, resulting in 17,651 total triplets spanning 15,602 distinct visual environments, representing over 500$\times$ more visual diversity compared to real data alone.

For video datasets used in training and evaluation, optical flow maps are computed using RAFT~\cite{RAFT} at original resolution to maintain data quality. Each video frame, consisting of an RGB image, optical flow map, and segmentation mask, is treated as an independent image-flow-mask triplet. This processing approach maintains consistency with the DUTSv2 structure and aligns with the per-frame processing paradigm adopted by modern unsupervised VOS methods, enabling seamless integration of both synthetic and real data during training.

\begin{table*}[t]
\centering 
\small
\caption{Quantitative evaluation on the DAVIS 2016 validation set and FBMS test set. OF and PP indicate the use of optical flow estimation model and post-processing technique, respectively. $\ast$ denotes speed calculated on a single GeForce RTX 2080 Ti GPU.}
\vspace{-2mm}
\begin{tabular}{p{2.2cm}cccP{6mm}P{6mm}P{6mm}P{8mm}P{8mm}P{8mm}P{8mm}}
\toprule
\multicolumn{7}{c}{} &\multicolumn{3}{c}{DAVIS 2016} &\multicolumn{1}{c}{FBMS}\\
\cmidrule(lr){8-11}
Method &Publication &Backbone &Resolution &OF &PP &fps &$\mathcal{G}_\mathcal{M}$ &$\mathcal{J}_\mathcal{M}$ &$\mathcal{F}_\mathcal{M}$ &$\mathcal{J}_\mathcal{M}$\\
\midrule
DFNet~\cite{DFNet} &ECCV'20 &DeepLabv3~\cite{deeplabv3} &- & &\checkmark &3.57 &82.6 &83.4 &81.8 &-\\
F2Net~\cite{F2Net} &AAAI'21 &DeepLabv3~\cite{deeplabv3} &473$\times$473 & & &10.0 &83.7 &83.1 &84.4 &77.5\\
RTNet~\cite{RTNet} &CVPR'21 &ResNet-101~\cite{resnet} &384$\times$672 &\checkmark &\checkmark &- &85.2 &85.6 &84.7 &-\\
FSNet~\cite{FSNet} &ICCV'21 &ResNet-50~\cite{resnet} &352$\times$352 &\checkmark &\checkmark &12.5 &83.3 &83.4 &83.1 &-\\
TransportNet~\cite{TransportNet} &ICCV'21 &ResNet-101~\cite{resnet} &512$\times$512 &\checkmark & &12.5 &84.8 &84.5 &85.0 &78.7\\
AMC-Net~\cite{AMC-Net} &ICCV'21 &ResNet-101~\cite{resnet} &384$\times$384 &\checkmark &\checkmark &17.5 &84.6 &84.5 &84.6 &76.5\\
D$^2$Conv3D~\cite{D2Conv3D} &WACV'22 &ir-CSN-152~\cite{csn} &480$\times$854 & & &- &86.0 &85.5 &86.5 &-\\
IMP~\cite{IMP} &AAAI'22 &ResNet-50~\cite{resnet} &- & & &1.79 &85.6 &84.5 &86.7 &77.5\\
HFAN~\cite{HFAN} &ECCV'22 &MiT-b2~\cite{mit} &512$\times$512 &\checkmark & &12.8$^\ast$ &87.5 &86.8 &88.2 &-\\
PMN~\cite{PMN} &WACV'23 &VGG-16~\cite{vgg} &352$\times$352 &\checkmark & &\underline{41.3}$^\ast$ &85.9 &85.4 &86.4 &77.7\\
TMO~\cite{TMO} &WACV'23 &ResNet-101~\cite{resnet} &384$\times$384 &\checkmark & &\textbf{43.2}$^\ast$ &86.1 &85.6 &86.6 &79.9\\
OAST~\cite{OAST} &ICCV'23 &MobileViT3D~\cite{mobilevit} &384$\times$640 &\checkmark & &- &87.0 &86.6 &87.4 &83.0\\
GFA~\cite{GFA} &AAAI'24 &- &512$\times$512 &\checkmark & &- &\underline{88.2} &\underline{87.4} &\underline{88.9} &82.4\\
GSA-Net~\cite{GSA-Net} &CVPR'24 &MiT-b2~\cite{mit} &512$\times$512 &\checkmark & &32.1$^\ast$ &\underline{88.2} &\underline{87.4} &\textbf{89.0} &82.3\\
DPA~\cite{DPA} &CVPR'24 &VGG-16~\cite{vgg} &512$\times$512 &\checkmark & &19.5$^\ast$ &87.6 &86.8 &88.4 &\underline{83.4}\\
\midrule
\textbf{DepthFlow} & &MiT-b2~\cite{mit} &512$\times$512 &\checkmark & &29.5$^\ast$ &\textbf{88.5} &\textbf{88.0} &\textbf{89.0} &\textbf{84.7}\\
\bottomrule
\end{tabular}
\label{table2}
\end{table*}

\begin{table*}[t]
\centering 
\small
\caption{Quantitative evaluation on the YouTube-Objects dataset. Performance is reported with the $\mathcal{J}$ mean.}
\vspace{-2mm}
\begin{tabular}{p{1.8cm}P{2.15cm}P{7.3mm}P{7.3mm}P{7.3mm}P{7.3mm}P{7.3mm}P{7.3mm}P{7.3mm}P{7.3mm}P{7.3mm}P{7.3mm}|P{7.3mm}}
\toprule
Method &Backbone &Aero. &Bird &Boat &Car &Cat &Cow &Dog &Horse &Motor. &Train &Mean\\
\midrule
COSNet~\cite{COSNet} &DeepLabv3~\cite{deeplabv3} &81.1 &75.7 &71.3 &77.6 &66.5 &69.8 &76.8 &67.4 &\underline{67.7} &46.8 &70.5\\
AGNN~\cite{AGNN} &DeepLabv3~\cite{deeplabv3} &71.1 &75.9 &70.7 &78.1 &67.9 &69.7 &77.4 &67.3 &\textbf{68.3} &47.8 &70.8\\
HFAN~\cite{HFAN} &MiT-b1~\cite{mit} &84.7 &80.0 &72.0 &76.1 &76.0 &71.2 &76.9 &\textbf{71.0} &64.3 &\underline{61.4} &73.4\\
TMO~\cite{TMO} &ResNet-101~\cite{resnet} &85.7 &80.0 &70.1 &78.0 &73.6 &70.3 &76.8 &66.2 &58.6 &47.0 &71.5\\
GFA~\cite{GFA} &- &87.2 &85.5 &\textbf{74.7} &\textbf{82.9} &80.4 &\underline{72.0} &79.6 &67.8 &61.3 &55.8 &\underline{74.7}\\
GSA-Net~\cite{GSA-Net} &MiT-b2~\cite{mit} &\textbf{87.9} &83.0 &69.5 &81.2 &77.9 &\textbf{74.2} &78.8 &\underline{68.7} &61.8 &59.3 &74.2\\
DPA~\cite{DPA} &VGG-16~\cite{vgg} &\underline{87.5} &\underline{85.6} &70.1 &77.7 &\textbf{81.2} &69.0 &\textbf{81.8} &61.9 &62.1 &51.3 &73.7\\
\midrule
\textbf{DepthFlow} &MiT-b2~\cite{mit} &86.9 &\textbf{86.2} &\underline{74.5} &\underline{82.5} &\underline{80.8} &71.8 &\underline{80.8} &66.1 &55.9 &\textbf{63.3} &\textbf{75.1}\\
\bottomrule
\end{tabular}
\label{table3}
\end{table*}

\subsection{Implementation Details}

\noindent\textbf{Network architecture.}
We adopt a simple encoder-decoder architecture similar to TMO~\cite{TMO} with two encoders processing RGB image and optical flow independently. Features from both encoders are fused at multiple layers through element-wise addition and decoded to produce segmentation masks. Following existing methods~\cite{BMVOS, TBD}, we incorporate CBAM~\cite{CBAM} attention modules after each fusion step.

\vspace{1mm}
\noindent\textbf{Network training.}
Our model is trained in an end-to-end manner using fully supervised learning with image-flow-mask triplets, learning to map input image-flow pairs directly to segmentation masks. We adopt a two-stage training protocol to provide comprehensive knowledge to the network. In the first stage, we pre-train the model using the YouTube-VOS 2018~\cite{YTVOS} training set, merging all annotated objects into single binary masks as salient objects. In the second stage, we fine-tune using a combination of DAVIS 2016 training set and our DUTSv2 dataset with a mixing ratio of 1:3, ensuring the network benefits from both real video data and our large-scale synthetic dataset. For optimization, we employ cross-entropy loss and Adam~\cite{adam} optimizer with learning rate 1e-5. All training is conducted on two GeForce RTX TITAN GPUs with standard data augmentation techniques.

\section{Experiments}
We conduct extensive experiments to validate the effectiveness of our proposed approach. For unsupervised VOS, we evaluate on DAVIS 2016~\cite{DAVIS} validation set, FBMS~\cite{FBMS} test set, YouTube-Objects~\cite{YTOBJ} dataset, and Long-Videos~\cite{LVID} dataset. For video SOD, we evaluate on DAVIS 2016 validation set, FBMS test set, DAVSOD~\cite{SSAV} test set, and ViSal~\cite{ViSal} dataset. We refer to our method as DepthFlow throughout this section for brevity.

\subsection{Qualitative Results}
Figure~\ref{figure5} showcases qualitative mask prediction results that demonstrate DepthFlow’s robustness in challenging scenarios. Despite severe occlusions, rapid camera motion, and complex background clutter, DepthFlow consistently segments primary objects with high accuracy. These results highlight the effectiveness of training with diverse synthetic flows to handle a broad spectrum of video complexities.

\subsection{Quantitative Results}

\noindent\textbf{Unsupervised video object segmentation.}
Tables~\ref{table2}, \ref{table3}, and \ref{table4} present quantitative comparisons between our method and existing state-of-the-art approaches on unsupervised VOS benchmarks. To ensure fair comparison, we specify the backbone architecture used in each method. The results demonstrate that constructing a robust training dataset is as crucial as designing sophisticated architectures. By improving only the training data through our depth-to-flow synthetic data generation, our method outperforms all existing approaches while maintaining efficient inference speeds. This validates that depth-to-flow conversion preserves the structural patterns necessary for effective VOS learning.

\vspace{1mm}
\noindent\textbf{Video salient object detection.}
Video SOD is closely related to unsupervised VOS and provides valuable evaluation of generalization capabilities. Table~\ref{table5} presents results on video SOD benchmarks, demonstrating consistent performance improvements across multiple datasets. These results further validate that depth-to-flow conversion is sufficient for learning robust models, as the preserved structural patterns enable strong performance across related video understanding tasks. The consistent improvements highlight the broad applicability of our approach.

\begin{table}[t!]
\centering 
\small
\caption{Quantitative evaluation on the Long-Videos dataset. SS and US denote semi-supervised and unsupervised, respectively.}
\vspace{-2mm}
\begin{tabular}{p{2.0cm}P{2.2cm}P{0.8cm}P{0.8cm}}
\toprule
Method &Backbone &Type &$\mathcal{J}_\mathcal{M}$\\
\midrule
STM~\cite{stm} &ResNet-50~\cite{resnet} &SS &\underline{79.1}\\
AFB-URR~\cite{LVID} &ResNet-50~\cite{resnet} &SS &\textbf{82.7}\\
\midrule
AGNN~\cite{AGNN} &DeepLabv3~\cite{deeplabv3} &US &68.3\\
HFAN~\cite{HFAN} &MiT-b2~\cite{mit} &US &80.2\\
GSA-Net~\cite{GSA-Net} &MiT-b2~\cite{mit} &US &\underline{80.6}\\
\midrule
\textbf{DepthFlow} &MiT-b2~\cite{mit} &US &\textbf{81.1}\\
\bottomrule
\end{tabular}
\label{table4}
\end{table}

\begin{table*}[t]
\centering 
\caption{Quantitative evaluation on the DAVIS 2016 validation set, FBMS test set, DAVSOD test set, and ViSal dataset.}
\vspace{-2mm}
\small
\begin{tabular}{p{2.0cm}|P{5.8mm}P{5.8mm}P{5.8mm}|P{5.8mm}P{5.8mm}P{5.8mm}|P{5.8mm}P{5.8mm}P{5.8mm}|P{5.8mm}P{5.8mm}P{5.8mm}|P{5.8mm}P{5.8mm}P{5.8mm}}
\toprule
\multirow{2}{*}{Method} & \multicolumn{3}{c|}{DAVIS 2016} & \multicolumn{3}{c|}{FBMS} & \multicolumn{3}{c|}{DAVSOD} & \multicolumn{3}{c|}{ViSal} & \multicolumn{3}{c}{Average} \\
 & $\mathcal{S}\uparrow$ & $\mathcal{F}\uparrow$ & $\mathcal{M}\downarrow$ & $\mathcal{S}\uparrow$ & $\mathcal{F}\uparrow$ & $\mathcal{M}\downarrow$ & $\mathcal{S}\uparrow$ & $\mathcal{F}\uparrow$ & $\mathcal{M}\downarrow$ & $\mathcal{S}\uparrow$ & $\mathcal{F}\uparrow$ & $\mathcal{M}\downarrow$  & $\mathcal{S}\uparrow$ & $\mathcal{F}\uparrow$ & $\mathcal{M}\downarrow$ \\
\midrule
SSAV~\cite{SSAV} &89.3 &86.1 &2.8 &87.9 &86.5 &4.0 &72.4 &60.3 &9.2 &94.3 &93.9 &2.0 &86.0 &81.7 &4.5\\
AD-Net~\cite{AD-Net} &- &80.8 &4.4 &- &81.2 &6.4 &- &- &- &- &90.4 &3.0 &- &- &-\\
F$^3$Net~\cite{F3Net} &85.0 &81.9 &4.1 &85.3 &81.9 &6.8 &68.9 &56.4  &11.7 &87.4 &90.7 &4.5 &81.7 &77.7 &6.8\\
MINet~\cite{MINet} &86.1 &83.5 &3.9 &84.9 &81.7 &6.7 &70.4 &58.2 &10.3 &90.3 &91.1 &4.1 &82.9 &78.6 &6.3\\
GateNet~\cite{GateNet} &86.9 &84.6 &3.6 &85.7 &83.2 &6.5 &70.1 &57.8 &10.4 &92.1 &92.8 &3.9 &83.7 &79.6 &6.1\\
PCSA~\cite{PCSA} &90.2 &88.0 &2.2 &86.6 &83.1 &4.1 &74.1 &65.5 &8.6 &94.6 &94.0 &1.7 &86.4 &82.7 &4.2\\
3DC-Seg~\cite{3DC-Seg} &- &91.8 &1.5 &- &84.5 &4.8 &- &- &- &- &92.2 &1.9 &- &- &-\\
CASNet~\cite{CASNet} &87.3 &86.0 &3.2 &85.6 &86.3 &5.6 &69.4 &- &8.9 &82.0 &84.7 &2.9 &81.1 &- &5.2\\
FSNet~\cite{FSNet} &92.0 &90.7 &2.0 &89.0 &88.8 &4.1 &77.3 &68.5 &\underline{7.2} &- &- &- &- &- &-\\
CFAM~\cite{CFAM} &91.8 &90.9 &1.5 &90.9 &91.5 &\underline{2.6} &75.3 &66.2 &8.3 &94.7 &\underline{95.1} &1.3 &88.2 &85.9 &3.4\\
UFO~\cite{UFO} &91.8 &90.6 &1.5 &89.1 &88.8 &3.1 &- &- &- &\textbf{95.9} &\underline{95.1} &1.3 &- &- &-\\
DBSNet~\cite{DBSNet} &92.4 &91.4 &1.4 &88.2 &88.5 &3.8 &77.8 &68.8 &7.6 &93.1 &92.8 &2.0 &87.9 &85.4 &3.7\\
HFAN~\cite{HFAN} &93.4 &\underline{92.9} &\textbf{0.9} &87.5 &84.9 &3.3 &75.3 &68.0 &\textbf{7.0} &94.1 &93.5 &\underline{1.1} &87.6 &84.8 &3.1\\
TMO~\cite{TMO} &92.8 &92.0 &\textbf{0.9} &88.6 &88.2 &3.1 &76.7 &70.8 &\underline{7.2} &94.2 &94.7 &\textbf{1.0} &88.1 &86.4 &3.1\\
OAST~\cite{OAST} &\underline{93.5} &92.6 &\underline{1.1} &\underline{91.7} &\underline{91.9} &\textbf{2.5} &\underline{78.6} &\underline{71.2} &\textbf{7.0} &94.8 &95.0 &\textbf{1.0} &\underline{89.7} &\underline{87.7} &\textbf{2.9}\\
\midrule
\textbf{DepthFlow} &\textbf{94.6} &\textbf{94.1} &\underline{1.1} &\textbf{92.2} &\textbf{92.0} &2.7 &\textbf{80.2} &\textbf{72.7} &\textbf{7.0} &\underline{95.7} &\textbf{96.3} &1.3 &\textbf{90.7} &\textbf{88.8} &\underline{3.0}\\
\bottomrule
\end{tabular}
\label{table5}
\end{table*}

\begin{table}[t]
\centering 
\small
\caption{Ablation study on the training data setting. Pre. indicates the use of pre-training with large-scale video segmentation data.}
\vspace{-2mm}
\begin{tabular}{c|c|c|P{5mm}P{5mm}P{5mm}P{5mm}}
\toprule
Version &Pre. &Training &D &F &Y &L\\
\midrule
\Romannum{1} & &Real &83.2 &66.2 &62.2 &60.5\\
\Romannum{2} & &Synthetic &79.5 &79.8 &74.9 &74.9\\
\Romannum{3} & &Mixed &87.6 &82.7 &72.5 &77.2\\
\midrule
\Romannum{4} &\checkmark &- &82.2 &81.7 &73.3 &68.3\\
\Romannum{5} &\checkmark &Real &87.7 &81.8 &72.8 &73.2\\
\Romannum{6} &\checkmark &Synthetic &85.8 &83.4 &76.5 &82.0\\
\Romannum{7} &\checkmark &Mixed &88.5 &84.7 &75.1 &81.1\\
\bottomrule
\end{tabular}
\label{table6}
\end{table}

\subsection{Analysis}
\noindent\textbf{Training protocol.}
Table~\ref{table6} compares different training protocols: real (DAVIS 2016~\cite{DAVIS} training set), synthetic (our DUTSv2 dataset), and mixed (both combined). Incorporating synthetic data significantly improves performance with and without pre-training. The gains are particularly substantial on FBMS~\cite{FBMS}, YouTube-Objects~\cite{YTOBJ}, and Long-Videos~\cite{LVID} datasets, which exhibit different domain characteristics from DAVIS. Remarkably, training exclusively on synthetic data outperforms real DAVIS training across these benchmarks, demonstrating that diverse synthetic training data provide more robust training signals than limited real video data.

\begin{table}[t]
\centering 
\small
\caption{Ablation study on the backbone network.}
\vspace{-2mm}
\begin{tabular}{P{1.5cm}|P{6mm}P{6mm}P{6mm}P{6mm}P{6mm}}
\toprule
Backbone &fps &D &F &Y &L\\
\midrule
MiT-b0 &61.1 &86.7 &81.2 &70.4 &75.7\\
MiT-b1 &45.5 &87.3 &81.8 &73.5 &77.9\\
MiT-b2 &29.5 &88.5 &84.7 &75.1 &81.1\\
\bottomrule
\end{tabular}
\label{table7}
\end{table}

\vspace{1mm}
\noindent\textbf{Backbone network.}
Table~\ref{table7} compares various backbone architectures. Larger backbones achieve higher performance across all datasets at the cost of increased computational requirements. The consistent performance improvements with larger backbones validate the quality of our synthetic data, as poor-quality data would limit model scaling regardless of architecture capacity. We use MiT-b2 as the default backbone for all experiments.

\vspace{1mm}
\noindent\textbf{Aligned comparison.}
To validate DepthFlow's effectiveness, Table~\ref{table8} presents direct comparisons with HFAN~\cite{HFAN}, which employs a similar two-stream pipeline and per-frame inference protocol. Our method consistently outperforms HFAN in both inference speed and segmentation accuracy across all backbone variants. Notably, with smaller backbones where pre-trained knowledge is less leveraged, HFAN struggles to extract meaningful information from limited high-quality training data, leading to overfitting. These results demonstrate that well-designed training data can be more impactful than sophisticated network architectures, supporting our core contribution.

\begin{table}[t]
\centering 
\small
\caption{Aligned comparison on the DAVIS 2016 validation set. The speed is calculated on the same hardware.}
\vspace{-2mm}
\begin{tabular}{p{1.5cm}|P{1.3cm}|P{6mm}P{6mm}P{6mm}P{6mm}}
\toprule
Method &Backbone &fps &$\mathcal{G}_\mathcal{M}$ &$\mathcal{J}_\mathcal{M}$ &$\mathcal{F}_\mathcal{M}$\\
\midrule
\multirow{3}*{HFAN~\cite{HFAN}} &MiT-b0 &21.8 &81.2 &81.5 &80.8\\
&MiT-b1 &18.4 &86.7 &86.2 &87.1\\
&MiT-b2 &12.8 &87.5 &86.8 &88.2\\
\midrule
\multirow{3}*{DepthFlow} &MiT-b0 &61.1 &86.7 &86.5 &87.0\\
&MiT-b1 &45.5 &87.3 &87.0 &87.6\\
&MiT-b2 &29.5 &88.5 &88.0 &89.0\\
\bottomrule
\end{tabular}
\label{table8}
\end{table}

\begin{table}[t]
\centering 
\small
\caption{Controlled experiment to evaluate model flexibility.}
\vspace{-2mm}
\begin{tabular}{p{1.5cm}|P{7mm}|P{6mm}P{6mm}P{6mm}P{6mm}}
\toprule
Method &Flow &D &F &Y &L\\
\midrule
\multirow{2}*{TMO~\cite{TMO}} & &80.0 &80.0 &73.1 &67.7\\
&\checkmark &86.1 &79.9 &71.5 &72.5\\
\midrule
\multirow{2}*{DepthFlow} & &86.2 &81.2 &75.3 &78.3\\
&\checkmark &88.5 &84.7 &75.1 &81.1\\
\bottomrule
\end{tabular}
\label{table9}
\end{table}

\begin{figure*}[t]
\centering
\includegraphics[width=1\linewidth]{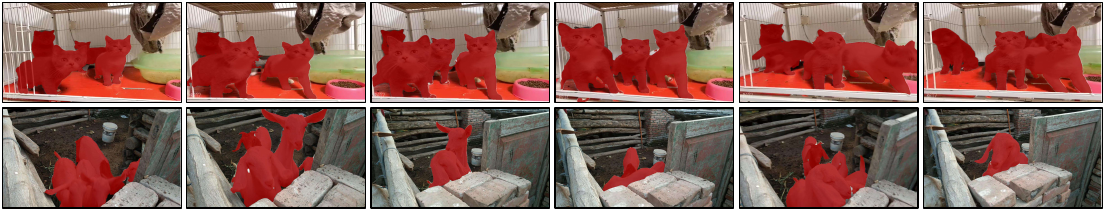}
\vspace{-6mm}
\caption{Qualitative results on the general video cases.}
\label{figure6}
\end{figure*}

\vspace{1mm}
\noindent\textbf{Model flexibility.}
While our method requires image-flow pairs for prediction, it can generate synthetic flows from individual video frames, enabling operation without real optical flows and effectively functioning as an image-level SOD model. To evaluate this flexibility, we compare with TMO~\cite{TMO}, which also provides motion-independent inference capability, as shown in Table~\ref{table9}. DepthFlow outperforms TMO across all benchmark datasets both with and without optical flow maps, demonstrating superior adaptability. Although using synthetic flows introduces minor performance variations compared to real flows, these differences remain negligible, confirming the robustness.

\vspace{1mm}
\noindent\textbf{General video cases.}
Figure~\ref{figure6} presents results on general videos from the MOSE~\cite{mose} validation set. Without any additional training or fine-tuning, the model trained under our setting consistently detects the salient objects, demonstrating strong generalization to real-world video scenarios.

\begin{figure}[t]
\centering
\includegraphics[width=1\linewidth]{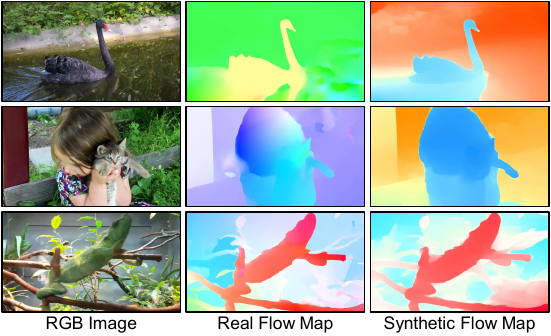}
\vspace{-6mm}
\caption{Visual comparison between real flow maps from flow estimation model and our generated flow maps.}
\label{figure7}
\end{figure}

\vspace{1mm}
\noindent\textbf{Comparison to real flows.}
Figure~\ref{figure7} provides qualitative comparison between real optical flows (estimated using pre-trained models) and our synthetic flows (generated via depth-to-flow conversion) on DAVIS 2017~\cite{DAVIS17} dataset. While the synthetic flows may not capture geometrically precise motion, they exhibit plausible motion patterns that effectively distinguish foreground from background. Our depth-to-flow conversion preserves essential structural patterns, demonstrating that structural motion information is sufficient for robust VOS training.

\begin{figure}[t]
\centering
\includegraphics[width=1\linewidth]{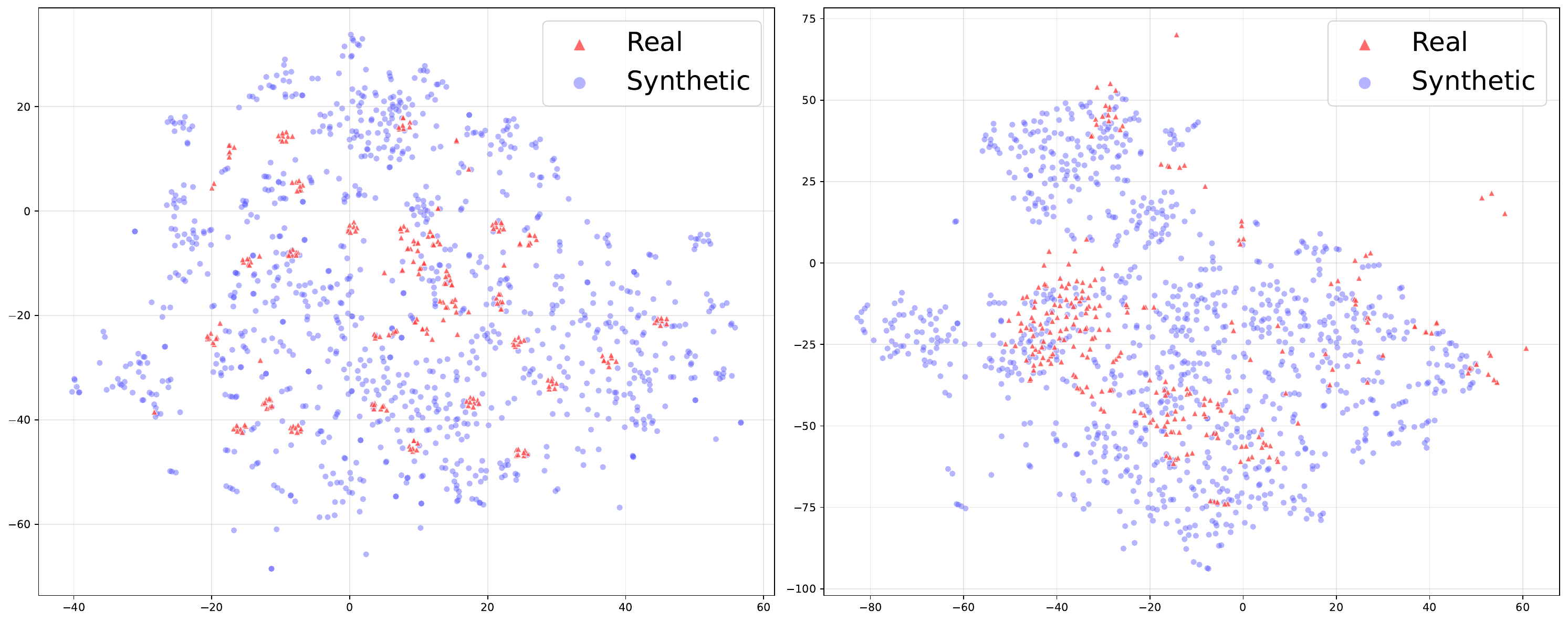}
\vspace{-6mm}
\caption{t-SNE comparison of feature distributions from real and synthetic datasets (left: RGB images, right: optical flow maps).}
\label{figure8}
\end{figure}

\vspace{1mm}
\noindent\textbf{Data distribution.}
In Figure~\ref{figure8}, we extract features from both RGB images and flow maps using pre-trained encoders and visualize their distributions using t-SNE. The analysis reveals that features from our synthetic DUTSv2 dataset closely align with those from real DAVIS 2016 data in both modalities. Importantly, the synthetic data also fills underrepresented regions in the feature space, effectively expanding the coverage of real data distributions. This demonstrates that our synthetic data generation approach not only maintains realistic feature characteristics but also provides enhanced diversity, supporting its effectiveness as a scalable training source for robust VOS models.

\vspace{1mm}
\noindent\textbf{Limitations.}
DepthFlow relies on predicted depth maps to generate synthetic training data, making the quality of the training data dependent on the accuracy of depth estimation. Inaccuracies in depth prediction, especially in challenging scenes, can lead to suboptimal flow generation and affect downstream performance. While our current framework uses real optical flow during inference, a promising future direction is to investigate whether depth-to-flow simulation can also help compensate for low-quality or unreliable flow estimates at test time, potentially offering a more robust motion representation under adverse conditions.

\section{Conclusion}
Data scarcity remains a key bottleneck in unsupervised VOS. We propose a depth-to-flow synthetic data generation method that achieves state-of-the-art performance without complex architectures or heavy post-processing. Our findings demonstrate that synthetic flows, when preserving structural information, can effectively complement real training data, showing that high-quality data is as important as network design. This work provides a strong foundation for future synthetic data generation in video understanding.

{
    \small
    \bibliographystyle{ieeenat_fullname}
    \bibliography{DepthFlow}
}

\end{document}